\documentclass[runningheads]{llncs}
\usepackage{graphicx}
\usepackage{amsmath,amssymb} 
\usepackage{color}
\usepackage{navigator}
\usepackage{multirow}
\usepackage[width=122mm,left=12mm,paperwidth=146mm,height=193mm,top=12mm,paperheight=217mm]{geometry}
\usepackage{epsfig}
\usepackage{gensymb}
\usepackage{caption}
\usepackage{subcaption}
\usepackage[bookmarks=false]{hyperref}
\usepackage{xcolor}
\hypersetup{
	colorlinks,
	linkcolor={red!70!black},
	citecolor={blue!50!black},
	urlcolor={blue!80!black}
}

\begin{document}
\pagestyle{headings}
\mainmatter

\title{Generic 3D Representation \\ via Pose Estimation and Matching}	

\titlerunning{Generic 3D Representation}


\authorrunning{Zamir, Wekel, Agarwal, Wei, Malik, Savarese}
\author{Amir R. Zamir$^{1,2}$ \;\; Tilman Wekel$^{1}$ \;\; Pulkit Agrawal$^{2}$ \;\;  Colin Wei$^{1}$ \;\; \\ \vspace{3pt} Jitendra Malik$^{2}$ \;\; Silvio Savarese$^{1}$}


\institute{$^1$ Stanford University \;\; $^2$ University of California, Berkeley  \\ \vspace{5pt}
\textcolor{blue}{\url{http://3Drepresentation.stanford.edu/}\vspace{-0pt}}
}

\newcommand{\smallsec}[1]{\vspace{0.1in} \noindent {\bf #1.}}
\newcommand{\todo}[1]{\textcolor{blue}{\textbf{(#1)}}}

\maketitle


\begin{abstract}

Though a large body of computer vision research has investigated developing generic semantic representations, efforts towards developing a similar representation for 3D has been limited. 
In this paper, we learn a generic 3D representation through solving a set of foundational proxy 3D tasks: object-centric camera pose estimation and wide baseline feature matching. Our method is based upon the premise that by providing supervision over a set of carefully selected foundational tasks, generalization to novel tasks and abstraction capabilities can be achieved. We empirically show that the internal representation of a multi-task ConvNet trained to solve the above core problems generalizes to novel 3D tasks (e.g., scene layout estimation, object pose estimation, surface normal estimation) without the need for fine-tuning and shows traits of abstraction abilities (e.g., cross modality pose estimation).

 In the context of the core supervised tasks, we demonstrate our representation achieves state-of-the-art wide baseline feature matching results without requiring apriori rectification (unlike SIFT and the majority of learnt features). We also show 6DOF camera pose estimation given a pair local image patches. The accuracy of both supervised tasks come comparable to humans. Finally, we contribute a large-scale dataset composed of object-centric street view scenes along with point correspondences and camera pose information, and conclude with a discussion on the learned representation and open research questions.

\keywords{Generic Vision, Representation, Descriptor Learning, Pose Estimation, Wide-Baseline Matching, Street View.}

\end{abstract}


\section{Introduction}
\label{sec:intro}
Supposed an image is given and we are interested in extracting some 3D information from it, such as, the scene layout or the pose of the visible objects. 
One potential approach would be to annotate a dataset for every single desired problem and train a fully supervised system for each (i.e., supervised learning). 
This is undesirable as an annotated dataset for each problem would be needed as well as the fact that the problems would be treated independently. In addition, unlike semantic annotations such as, object labels, certain annotations in 3D are cumbersome to collect and often require special sensors (imagine manually annotating exact pose of an object or surface normals). 
An alternative approach is to develop a system with a rather generic perception that can conveniently generalize to novel tasks. In this paper, we take a step towards developing a generic 3D perception system that 1) can solve novel 3D problems without fine-tuning, and 2) is capable of certain abstract generalizations in the 3D context (e.g., reason about pose similarity between two drastically different objects).

But, how could one learn such a generalizable system? Cognitive studies suggest living organisms can perform cognitive tasks for which they have not received supervision by supervised learning of other foundational tasks~\cite{twokitten,smith2005development,rader1980nature}.  Learning the relationship between visual appearance and changing the vantage point (self-motion) is among the first visual skills developed by infants and play a fundamental role in developing other skills, e.g., depth perception. 
A classic experiment~\cite{twokitten} showed a kitten that was deprived from self-motion experienced fundamental issues in 3D perception, such as failing to understand depth when placed on the Visual Cliff~\cite{visualcliff}. Later works~\cite{rader1980nature} argued this finding was not, at least fully, due to motion intentionality and the supervision signal of self-motion was indeed a crucial elements in learning basic visual skills.
What these studies essentially suggest are: 1) by receiving supervision on a certain proxy task (in this case, self-motion), other tasks (depth understanding) can be solved sufficiently without requiring an explicit supervision, 2) some vision tasks are more foundational than others (e.g., self-motion perception vs depth understanding).

			\begin{figure}
				\centering
				\includegraphics[width=1\columnwidth]{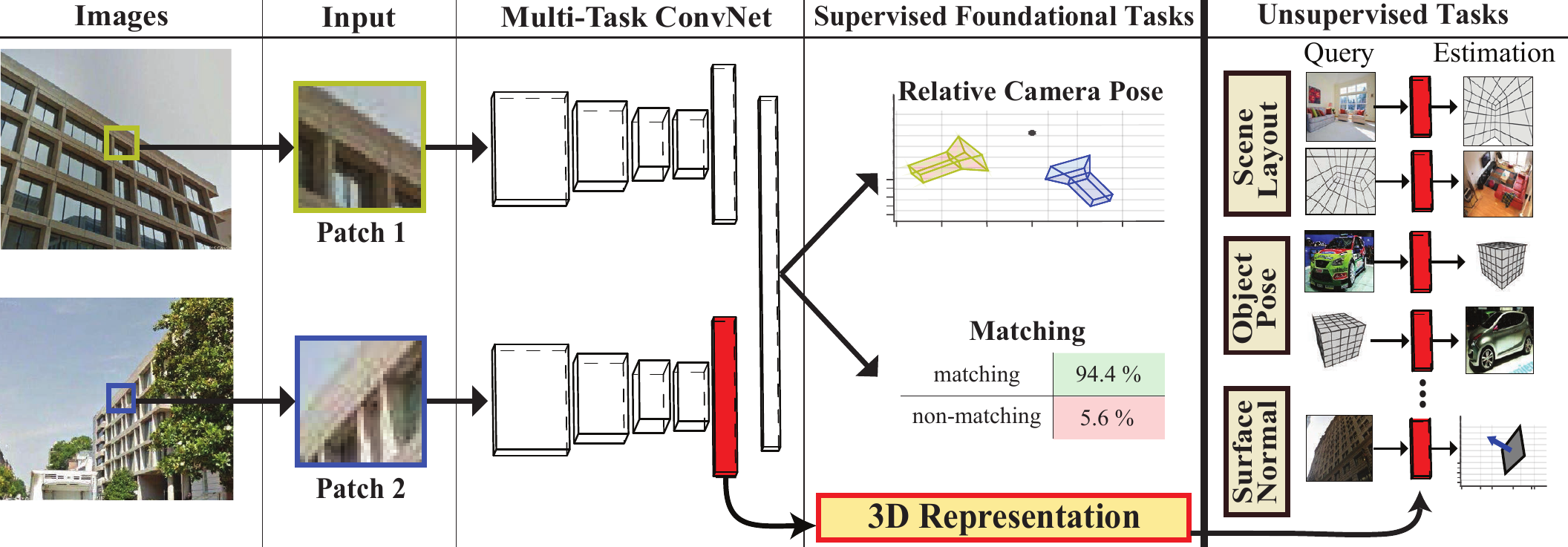}
				\caption{\scriptsize{\textbf{Learning a generic 3D representation}: we develop a supervised joint framework for camera pose estimation and wide baseline matching. We then show the internal representation of this framework can be used as a 3D representation generalizable to various 3D prediction tasks.\vspace{0pt}}}
				\label{fig:teaser}
			\end{figure} 

Inspired by the above discussion, we develop a supervised framework where a ConvNet is trained to perform 6DOF camera pose estimation. This basic task allows learning the relationship between an arbitrary change in the viewpoint and the appearance of an object/scene-point. One property of our approach is performing the camera pose estimation in a object/scene-centric manner: the training data is formed of image bundles that show the \emph{same point of an object/scene} while the camera moves around (i.e., it fixates - see the figure \ref{fig:datacol} (c)). This is different from existing video+metadata datasets~\cite{kitti}, the problem of Visual Odometry~\cite{nister2004visual,kitti}, and recent works on ego-motion estimation~\cite{LSM2015,dinesh}, where in the training data, the camera moves independent of the scene. Our object/scene-centric approach is equivalent to allowing a learner to focus on a physical point while moving around and observing how the appearance of that particular point transforms according to viewpoint change. Therefore, the learner receives an additional piece of information that the observed pixels are indeed showing the same object, giving more information about how the element looks under different viewpoints and providing better grounds for learning visual encoding of an observation. 
Infants also explore object-motion relationships~\cite{smith2005development} in a similar way as they hold an object in hand and observe it from different views. 

Our dataset also provides supervision for the task of wide baseline matching, defined as identifying if two images/patches are showing the same point regardless of the magnitude of viewpoint change. Wide baseline matching is also an important 3D problem and is closely related to object/scene-centric camera pose estimation: to identify whether two images could be showing the same point despite drastic changes in the appearance, an agent could learnt how viewpoint change impacts the appearance. Therefore, we perform our supervised training in a multi-task manner to simultaneously solve for both wide baseline matching and pose estimation.
This has the advantage of learning a single representation that encodes both problems. In experiments section (\ref{sec:joint}), we show it is possible to have a single representation solving both problems without a performance drop compared to having two dedicate representations. This provides practical computational and storage advantages. Also, training ConvNets using multiple tasks/losses is desirable as it has been shown to be better regularized~\cite{Xu2015,girshick15fastrcnn,weston2012deep}.\footnote{Though visual matching/tracking is also one of early developed cognitive skills\cite{infanttracking}, we are unware of any studies investigating its foundational role in developing visual perception. Therefore, we presume (and empirically observe) that the generality of our 3D representation is mostly attributed to the camera pose estimation component.} 

We train the ConvNet (siamese structure with weight sharing) on patch pairs extracted from the training data and use the last FC vector of one siamese tower as the generic 3D representation (see figure~\ref{fig:teaser}). We will empirically investigate if this representation can be used for solving novel 3D problems (we evaluated on scene layout estimation, object pose estimation, surface normal estimation), and whether it can perform any 3D abstraction (we experimented on cross category pose estimation and relating the pose of synthetic geometric elements to images).

\textbf{Dataset}:
We developed an object-centric dataset of street view scenes from the cities of Washington DC, NYC, San Francisco, Paris, Amsterdam, Las Vegas, and Chicago, augmented with \emph{camera pose} information and \emph{point correspondences}. It includes 25 million images, 118 million matching image pairs, camera metadata, 3D models of 8 cities. We release the dataset, trained models, and an online demo at \textcolor{blue}{\url{http://3Drepresentation.stanford.edu/}}.

\textbf{Novelty in the Supervised Tasks}:
Independent of providing a generic 3D representation, our approach to solving the two supervised tasks is novel in a few aspects. There is a large amount of previous work on detecting, describing, and matching image features, either through a handcrafting the feature~\cite{sift,surf,mikola,mser,alahi2012freak,pernoa_survey} or learning it~\cite{simonyan,matchnet,cvpr15,balntas2016pn,zbontar2015computing,simo2015discriminative,cnnunsup}. Unlike the majority of such features that utilize pre-rectification (within either the method or the training data), we argue that rectification prior to descriptor matching is not required; our representation can learn the impact of viewpoint change, rather than canceling it (by directly training on non-rectified data and supplying camera pose information during training). Therefore, it does not need an apriori rectification and is capable of performing wide baseline matching at the descriptor level. We report state-of-the-art results on feature matching. 
Wide baseline matching has been also the topic of many papers~\cite{pritchett1998wide,tell2002combining,goedeme2004fast,xiao2003two,zhang2012tilt} with the majority of them focused on leveraging various geometric constraints for ruling out incorrect `already-established' correspondences, as well as a number of methods that operate based on generating exhaustive warps~\cite{morel2009asift} or assuming 3D information about the scene is given~\cite{wu20083d}. In contrast, we learn a descriptor that is supervised to internally handle a wide baseline in the first place.

In the context of pose estimation, we show estimating a 6DOF camera pose given only a pair of local image patches, and without the need for several point correspondences, is feasible. 
This is different from many previous works \cite{rome,zissermanBook,visualsfm,song2013parallel,badino2013visual,chen2011city,kitti} from both visual odometery and SfM literature that perform the estimation through a two step process consisting of finding point correspondences between images followed by pose estimation. Koser and Koch~\cite{koser2008differential} also demonstrate pose estimation from a local region, though the plane on which the region lies is assumed to be given. The recent works of~\cite{LSM2015,dinesh} supervise a ConvNet on the camera pose from image batches but do not provide results on matching and pose estimation.  We report a human-level accuracy on this task.

\textbf{Existing Unsupervised Learning and ConvNet Initialization Works}:
The majority of previous unsupervised learning, transfer learning, and representation learning works have been targeted towards semantics~\cite{decaf,razavian2014cnn,sermanet2013overfeat,doersch2015unsupervised,tarr1994computational}. It has been practically well observed~\cite{decaf,razavian2014cnn} that the representation of a convnet trained on imagenet~\cite{krizhevsky2012imagenet} can generalize to other, mostly semantic, tasks. A number of methods investigated initialization techniques for ConvNet training based on unsupervised/weakly supervised data to alleviate the need for a large training dataset for various tasks~\cite{cmuvideo,doersch2015unsupervised}. Very recently, the methods of~\cite{LSM2015,dinesh} explored using motion metadata associated with videos (KITTI dataset~\cite{kitti}) as a form of supervision for training a ConvNet. However, they either do not investigate developing a 3D representation or intent to provide initialization strategies that are meant to be fine-tuned with supervised data for a desired task. In contrast, we investigate developing a generalizable 3D representation, perform the learning in an object-centric manner, and evaluate its unsupervised performance on various 3D tasks without any fine-tuning on the representation. We experimentally compare against the related recent works that made their models available~\cite{LSM2015,cmuvideo}.

Primary contributions of this paper are summarized as:
I) A generic 3D representation with empirically validated abstraction and generalization abilities. 
II) A learned joint descriptor for wide baseline matching and camera pose estimationat at the level of local image patches. 
III) A large-scale object-centric dataset of street view scenes including camera pose and correspondence information.

\section{Object-Centric Street View Dataset}
\label{sec:data_collection}
The dataset for the formulated task needs to not only provide a large amount of training data, but also show a rich camera pose variety, while the scale of the aimed learning problem invalidates any manual procedure.
We present a procedure that allows acquiring a large amount of training data in an automated manner, based on two sources of information: 1) Google street view~\cite{streetview} which is an almost inexhaustible source of geo-referenced and calibrated images, 2) 3D city models~\cite{dc3dmodels,streetview} that cover thousands of cities around the world. 
The dataset including 25 million images and 118 million matching image pairs with their camera pose and 3D models of 8 cities is available {\href{https://github.com/amir32002/3D_Street_View}{here}}.

\begin{figure}
	\includegraphics[width=1\textwidth]{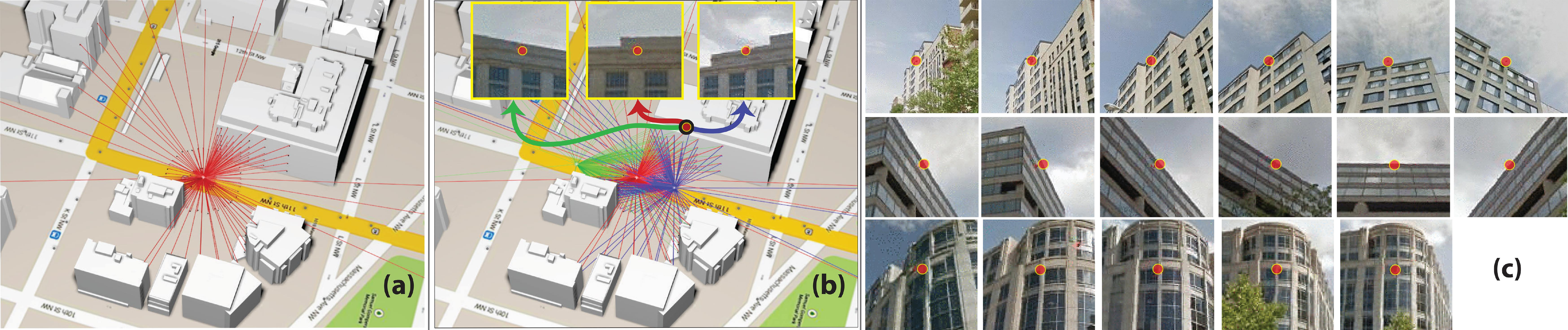}
	\caption{\scriptsize{\textbf{Illustration of the object-centric data collection process.} We use large-scale geo-registered 3D building models to register pixels in street view images on world coordinates system (see (a)) and use that for finding correspondences and their relative pose across multiple street view images (see (b)). Each ray represents one pixel-3D world coordinate correspondence. Each of the red, green, and blue colors represent one street view location. Each row in (c) shows a sample collected image bundle. The center pixel (marker) is expected to correspond to the same physical point.\vspace{0pt}}}
	\label{fig:datacol}
\end{figure}

The core idea of our approach is to form correspondences between the geo-referenced street view camera and physical 3D points that are given by the 3D models. More specifically, at any given street view location, we densely shoot rays into space in order to find intersections with nearby buildings. Each ray back projects one image pixel into the 3D space, as shown in figure~\ref{fig:datacol}-(a). By projecting the resulting intersection points onto adjacent street view panoramas (see figure~\ref{fig:datacol}-b), we can form image to image correspondences (see figure~\ref{fig:datacol}c). Each image is then associated with a (virtual) camera that fixates on the physical target point on a building by placing it on the optical center. To make the ray intersection procedure scalable, we perform occlusion reasoning on the 3D models to pre-identify from what GPS locations an arbitrary target would be visible and perform the ray intersection on those points only.

\textbf{Pixel Alignment and Pruning:}
This system requires integration of multiple resources, including elevation maps, GPS from street view, and 3D models. Though the quality of output exceeded our expectation (see samples in Figure~\ref{fig:datacol} (c)), any slight inaccuracy in the metadata or 3D models can cause a pixel misalignment in the collected images (examples shown in the first and last rows of figure~\ref{fig:datacol} (c)). Also, there are undocumented objects such as trees or moving objects that cause occlusions. Thus, a content-based post alignment and pruning was necessary. We again used metadata in our alignment procedure to be able to handle image bundles with arbitrarily wide baselines (note that the collected image bundles can show large, often$>100\degree$, viewpoint changes). In the interest of space, we describe this procedure in {\href{http://3drepresentation.stanford.edu/supplementary_material}{supplementary material}} (section~3).

This process forms our dataset composed of matching and non-matching patches as well as the relative camera pose for the matching pairs. We stopped collecting data when we reached the coverage of $>200{km}^2$ from the 7 cities mentioned in section~\ref{sec:intro}. The collection procedure is currently done on Google street view, but can be performed using any geo-referenced calibrated imagery.
We will experimentally show that the trained representation on this data does not manifest a clear bias towards street view scenes and outperforms existing feature learning methods on non-street view benchmarks.

\textbf{Noise Statistics:}
We performed a user study through Amazon Mechanical Turk to quantify the amount of noise in the final dataset. Please see {\href{http://3drepresentation.stanford.edu/supplementary_material}{supplementary material}} (section 3.2) for the complete discussion and results. Briefly, $68\%$ of the patch pairs were found to have at least $25\%$ of overlap in their content. The mean and standard deviation of pixel misalignment was 16.12 ($\approx11\%$ of patch width) and 11.55 pixels, respectively. We did not perform any filtering or geo-fencing on top of the collected data as the amount of noise appeared to be within the robustness tolerance of ConvNet trainings and they converged.

\section{Learning using ConvNets}
\label{sec:results}

A joint feature descriptor was learnt by supervising a Convolutional Neural Network (ConvNet) to perform 6DOF camera pose estimation and wide baseline matching between pairs of image patches. For the purpose of training, any two image patches depicting the same physical target point in the street view dataset were labelled as matching and other pairs of images were labelled as non-matching. The training for camera pose estimation was performed using matching patches. The patches were always cropped from the center of the collected street view image to keep the optical center at the target point.

The camera pose between each pair of matching patches was represented by a 6D vector; the first three dimensions were Tait-Bryan angles (roll, yaw, pitch) and the last three dimensions were cartesian (x, y, z) translation coordinates expressed in meters. For the purpose of training, 6D pose vectors were pre-processed to be zero mean and unit standard deviation (i.e., z-scoring). The ground-truth and predicted pose vectors for the $i^{th}$ example are denoted by $p^{*}_i, ~p_i$ respectively. The pose estimation loss $L_{pose}(p^{*}_i, p_i)$ was set to be the robust regression loss described in equation \ref{eq:robust}: 
\begin{equation}
\label{eq:robust}
L_{pose}(p^{*}_i, p_i) = 
    \left\{
	\begin{array}{ll}
		e & \mbox{if } e \leq 1 \\
		1 + \log e & \mbox{if } e > 1 
	\end{array}
\right.
\mbox{   where   }   e={||p^{*}_i-p_i||}_{l_2}.
\end{equation}
The loss function for patch matching $L_{match}({m_i^{*}, m_i})$ was set to be sigmoid cross entropy, where $m_i^{*}$ is the ground-truth binary variable indicating matching/non-matching and $m_i$ is the predicted probability of matching. 

ConvNet training was performed to optimize the joint matching and pose estimation loss ($L_{joint}$) described in equation \ref{eq:joint}. The relative weighting between the pose ($L_{pose}$) and matching ($L_{match}$) losses was controlled by $\lambda$ (we set $\lambda = 1$). 
\begin{equation}
\label{eq:joint}
    L_{joint}(p^{*}_i, m_i^{*}, p_i, m_i) = L_{pose}(p^{*}_i, p_i) + \lambda L_{match}(m_i, m^{*}_i).
\end{equation}

Our training set consisted of patch pairs drawn from a wide distribution of baseline changes ranging from $0\degree$ to over $120\degree$. We consider patches of size 192x192 ($<15\%$of the actual image size) and rescaled them to 101x101 before passing them into the ConvNet.


A ConvNet model with siamese architecture~\cite{chopra2005learning} containing two identical streams with identical set of weights was used for computing the relative pose and the matching score between the two input patches. A standard ConvNet architecture was used for each stream: C(20, 7, 1)-ReLU-P(2, 2)-C(40, 5, 1)-ReLU-P(2, 2)-C(80, 4, 1)-ReLU-P(2, 2)-C(160, 4, 2)-ReLU-P(2, 2)-F(500)-ReLU-F(500)-ReLU. The naming convention is as follows: C($n, k, s$): convolutional layer $n$ filters, spatial size $k\times k$, and stride $s$. P($k, s$): max pooling layer of size $k\times k$ and stride $s$. ReLU: rectified linear unit. F($n$): fully connected linear layer with $n$ output units. The feature descriptors of both streams were concatenated and fed into a fully connected layer of 500 units which were then fed into the pose and matching losses. With this ConvNet configuration, the size of the image representation (i.e., the last FC vector of one siamese half - see Figure~\ref{fig:teaser}) is 500. Our architecture is admittedly pretty common and standard. This allows us to evaluate if our good end performance is attributed to our hypothesis on learning on foundational tasks and the new dataset, rather than a novel architecture.

We trained the ConvNet model from scratch (i.e., randomly initialized weights) using SGD with momentum (initial learning rate of .001 divided by 10 per 60K iterations), gradient clipping, and a batch size of 256. We found that the use of gradient clipping was essential for training as even robust regression losses produce unstable gradients at the starting of training. Our network converged after 210K iterations. Training using Euler angles performed better than quaternions ($17.7\degree$~vs~$29.8\degree$ median angular error), and the robust loss outperformed the non-robust $l_2$ loss ($17.7\degree$~vs~$22.3\degree$ median angular error). Additional details about the training procedure can be found in the {\href{http://3drepresentation.stanford.edu/supplementary_material}{supplementary material}}.

\section{Experimental Discussions and Results}
\label{sec:results}
We implemented our framework using data parallelism~\cite{krizhevsky2014one} on a cluster of 5-10 GPUs. 
At the test time, computing the representation is a feed-forward pass through a siamese half ConvNet and takes $\sim2.9ms$ per image on a single processor.
Sections~\ref{sec:supervised_eval} and~\ref{sec:unsupervised_eval} provide the evaluations of the learned representation on the supervised and novel 3D tasks, respectively.



\subsection{Evaluations on the Supervised Tasks}
\label{sec:supervised_eval}
\subsubsection{Evaluations on the Street View Dataset.}
\label{sec:results_pose}
 The test set of pose estimation is composed of 7725 pairs of matching patches from our dataset. The test set of matching includes 4223 matching and 18648 non-matching pairs. It is made sure that no data from those areas and their vicinity is used in training. Each patch pair in the test sets was verified by three Amazon Methanical Turkers to verify the ground truth is indeed correct. For the matching pairs, the Turkers also ensured the center pixel of patches are no more than 25 pixels ($\sim3\%$ of image width) apart. Visualizations of the test set can be seen on our {\href{http://3Drepresentation.stanford.edu/}{website}}.
 
  \begin{figure*}
  	\centering
  	\includegraphics[width=1\textwidth]{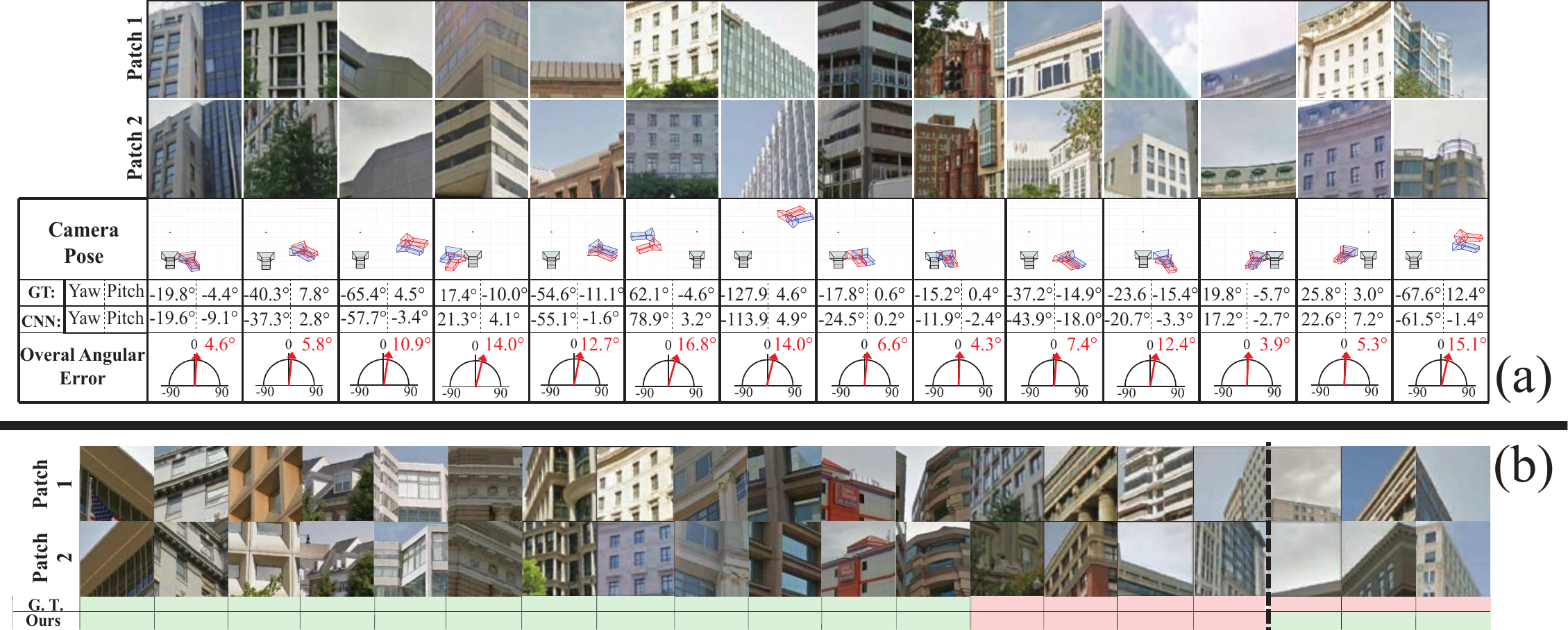}\vspace{0pt}
  	\caption{\scriptsize{\textbf{(a) Sample qualitative results of camera pose estimation}. $1^{st}$ and $2^{nd}$ rows show the patches. The $3^{rd}$ row depicts the estimated relative camera poses on a unit sphere (black: patch 1's camera (reference), red: ground-truth pose of patch 2, blue: estimated pose of patch 2). Rightward and upward are the positive directions. \textbf{(b)Sample wide baseline matching results.} Green and red represent `matching' and `non-matching', respectively. Three failure cases are shown on the right.}\vspace{-30pt}}
  	\label{fig:pose_ver_qual}
  \end{figure*}
  

\subsubsection{Pose Estimation.}
Figure~\ref{fig:pose_ver_qual}-(a) provides qualitative results of pose estimation. The angular evaluation metric is the standard overall angular error~\cite{kummerle2009measuring,kitti}, defined as the angle between the predicted pose vector and the ground truth vector in the plane defined by their cross product. The translational error metric is $l_2$ norm of the difference vector between the normalized predicted translation vector and ground truth~\cite{kitti,kummerle2009measuring}. The translation vector was normalized to enable comparing with up-to-scale SfM.

\begin{figure}
	\centering
	\includegraphics[width=1\columnwidth]{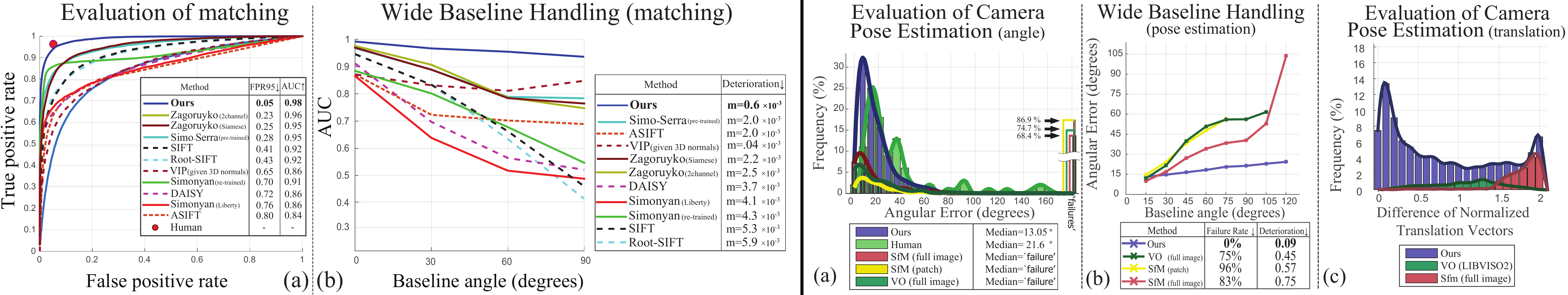}
	\caption{\scriptsize{\textbf{Left: Quantitative evaluation of matching.} ROC curves of each method and corresponding AUC and FPR@95 values are shown in (a). \textbf{Right: Quantitative evaluation of camera pose estimation.} VO and SfM denote Visual Odometery (LIBVISO2) and Structure-from-Motion (visualSfM), respectively. Evaluation of robustness to wide baseline camera shifts is shown in~(b)~plots.\vspace{0pt}}}
	\label{fig:quant_all}
\end{figure}

Figure~\ref{fig:quant_all}-right provides the quantitative evaluations. The plots (a) and (c) illustrate the distribution of the test set with respect to pose estimation error for each method (the more skewed to the left, the better). 
The green curve shows pose estimation results by human subjects. Two users with computer vision knowledge, but unaware of the particular use case, were asked to estimated the relative pitch and yaw between a random subset of 500 test pairs. They were allowed to train themselves with as many training sampled as they wished.
ConvNet outperformed human on this task with a margin of $8\degree$ in median error. 

\textbf{Pose Estimation Baselines:} We compared against Structure-from-Motion (visualSfM~\cite{visualsfm,wu2011multicore} with default components and tuned hyper-parameters for pairwise pose estimation on $192\times 192$ patches and full images) and LIBVISO2 Visual Odometery~\cite{Geiger2011IV} on full images. Both SfM and LIBVISO2 VO suffer from a large RANSAC failure rate mostly due to the wide baselines in test pairs.

Figure~\ref{fig:quant_all}-right (b) shows how the median angular error (Y axis) changes as the baseline of the test pairs (X axis) increases. This is achieved through binning the test set into 8 bins based on their baseline size. This plot quantifies the ability of the evaluated methods in handling a wide baseline. We adopt the slope of the curves as the quantification of deterioration in accuracy as the baseline increases. 


\subsubsection{Wide Baseline Matching.}
Figure~\ref{fig:pose_ver_qual}-(b) shows samples feature matching results using our approach, with three failure cases on the right. 
Figure~\ref{fig:quant_all}-left provides the quantitative results. The standard metric~\cite{brown} for descriptor matching is ROC curve acquired from sorting the test set pairs according to their matching score. For unsupervised methods, e.g., SIFT, the matching score is the $l_2$ distance. False Positive Rate at 95\% recall (FPR@95) and Area Under Curve (AUC) of ROC are standard scalar quantifications of descriptor matching~\cite{brown,simonyan}.

\textbf{Matching Baselines:} We compared our results with the handcrafted features of SIFT~\cite{sift}, Root-SIFT~\cite{rootsift}, DAISY~\cite{daisy}, VIP~\cite{wu20083d} (which requires the surface normals in the input for which we used the normals from the 3D models), and ASIFT~\cite{morel2009asift}. The matching score of ASIFT was the number of found correspondences in the test pair given the \emph{full images}. We also compared against the learning based features of Zagoruyko \& Komodakis~\cite{cvpr15} (using the models of authors), Simonyan \& Zisserman~\cite{simonyan} (with and without retraining), Simo-Serra et al.~\cite{simo2015discriminative} (using authors' best pretrained model) as well as human subjects (the red dot on the ROC plot). 
Figure~\ref{fig:quant_all}-left(b) provides the evaluations in terms of handling wide baselines, similar to Figure~\ref{fig:quant_all}-right(b).



\subsubsection{Brown et al. Benchmark \& Mikolajczyk's Benchmark.}
We performed evaluations on the \emph{non-street view} benchmarks of Brown et al.~\cite{brown} and Mikolajczyk \& Schmid~\cite{mikola} to find if 1) if our representation was performing well only on street view scenery, and 2) if wide baseline handling capability was achieved at the expense of lower performance on small baselines (as these benchmarks have a narrower baseline compared to our dataset for the most part). Tables~\ref{tab:brown}\&\ref{tab:Mikola} provide the quantitative results. We include a thorough description of evaluation setup and detailed discussions in the {\href{http://3drepresentation.stanford.edu/supplementary_material}{supplementary material}} (section~2).

\begin{table*}
	\label{tab:joint}
	\parbox{.58\linewidth}{
		\centering		
		\caption{\scriptsize{Evaluations on Brown's Benchmark~\cite{brown}. FPR@95 ($\downarrow$) is the metric.}}
		\label{tab:brown}
		\vspace{0pt}
		\scalebox{0.65}{
			\begin{tabular}
				{ @{}c@{} | @{}c@{} | @{}c@{} | @{}c@{} | @{}c@{} | @{}c@{} | @{}c@{} | @{}c@{} | @{}c@{} }
				{\bf Train } & {\bf{ Test }} & {  \scriptsize{ MatchNet }} & {{ \scriptsize{ Zagor. }}}& {{\scriptsize{ Simonyan~  }}}& {{\scriptsize{ Trzcinski }}} & {{ \scriptsize{ Brown }}} & {{ \scriptsize{ Root-SIFT }}} & {{\bf \scriptsize{ Ours}}}   \\ 
				{ } & { } & ~\cite{matchnet} & \scriptsize{siam~\cite{cvpr15}} & ~\cite{simonyan} & ~\cite{Trzcinski}  &  ~\cite{brown} &  ~\cite{rootsift}  & { }   \\ \hline
				Yos & ND  & 7.70 & 5.75 & 6.82 & 13.37 & 11.98 & 22.06 & \textbf{4.17} \\
				Yos & Lib  & 13.02 & 13.45 & 14.58 & 21.03 & 18.27 & 29.65 & \textbf{11.66} \\
				Lib & ND  & 4.75 & 4.33 & 7.22 & 14.15 & N/A & 22.06 & \textbf{1.47} \\ 
				ND & Lib  & 8.84 & 8.77 & 12.42 & 18.05 & 16.85 & 29.65 & \textbf{7.39} \\ \hline
				Lib & Yos  & 13.57 & 14.89 & \textbf{11.18} & 19.63 & N/A & 26.71 & 13.78 \\ 
				ND & Yos  & 11.00 & 13.23 & \textbf{10.08} & 15.86 & 13.55 & 26.71 & 12.30 \\\hline
				\multicolumn{2}{c|}{{mean}}  & 9.81  & 10.07 & 10.38 & 17.01 & 15.16  & 26.14 & \bfseries{8.46} \\
			\end{tabular}}
		}
		\hfill \hspace{0.01mm}
		\parbox{.42\linewidth}{
			\centering
			\vspace{0pt}
			\caption{\scriptsize{Evaluation on Mikolajczyk \& Schmid's~\cite{mikola}. The metric is mAP($\uparrow$).}}
			\label{tab:Mikola}
			\vspace{11pt}
			\scalebox{0.65}{
				\begin{tabular}{ c | c  c  c  c  c }
					\textbf{\footnotesize{Transf.}} & \multirow{2}{*}{\textbf{\small{1}}} & \multirow{2}{*}{\textbf{\small{2}}} & \multirow{2}{*}{\textbf{\small{3}}} & \multirow{2}{*}{\textbf{\small{4}}} & \multirow{2}{*}{\textbf{\small{5}}}\\
					\textbf{\footnotesize{Magnitude}} & & & & & \\ \hline
					\footnotesize{SIFT~\cite{sift}} &  40.1 &  28.0 & 24.3 & 29.0 &  17.1 \\
					\footnotesize{Zagor.~\cite{cvpr15}} & 43.2 & 37.5 & 29.2 & 28.0 & 16.8 \\ 
					\footnotesize{Fischer et al~\cite{cnnunsup}} & 42.3 & 33.9 & 26.1 & 22.1 & 14.6 \\ \hline
					\footnotesize{Ours-rectified} & 46.4 & \textbf{41.3} & 29.5 & 23.7 & 17.9 \\ 
					\footnotesize{Ours-unrectified} & \textbf{51.4} & 37.8 & \textbf{34.2} & \textbf{30.8} & \textbf{20.8} \\
				\end{tabular}}
			}
			\vspace{0pt}
		\end{table*}

\subsubsection{Joint Feature Learning.}
\label{sec:joint}
We studied different aspects of joint learning the representation and information sharing among the core supervised tasks. In the interest of space, we provide quantitative results in {\href{http://3drepresentation.stanford.edu/supplementary_material}{supplementary material}} (section 1). The conclusion of the tests was that: First, the problems of wide baseline matching and camera pose estimation have a great deal of shared information. Second, one descriptor can encode both problems with no performance drop.

\subsection{Evaluating the 3D Representation on Novel Tasks}
\label{sec:unsupervised_eval}
The results of evaluating our representation on novel 3D tasks are provided in this section. The tasks as well as the images (e.g., Airship images from ImageNet) used in these evaluations are significantly different from what our representation was trained for (i.e., camera pose estimation and matching on local patches of street view images). The fact that, despite such differences, our representation achieves best results among all unsupervised methods and gets close to supervised methods for each of the tasks empirically validates our hypothesis on learning on foundational tasks (see section~\ref{sec:intro}). 

Our ways of evaluating and probing the representation in an unsupervised manner are 1) tSNE~\cite{tsne}: large-scale 2D embedding of the representation. This allows visualizing the space and getting a sense of similarity from the perspective of the representation, 2) Nearest Neighbors (NN) on the full dimensional representation, and 3) training a simple classifier (e.g., KNN or a linear classifier) on the \emph{frozen}  representation (\underline{ i.e., no fine-tuning}) to read out a desired variable. The latter enables quantifying if the required information for solving a novel task is encoded in the representation and can be extracted using a simple function. We compare against the representations of related methods that made their models available~\cite{LSM2015,cmuvideo}, various layers of AlexNet trained on ImageNet~\cite{krizhevsky2012imagenet}, and a number of supervised techniques for some of the tasks. Additional results are provided in the {\href{http://3drepresentation.stanford.edu/supplementary_material}{supplementary material}} and the {\href{http://3Drepresentation.stanford.edu/}{website}}.

\begin{figure*}
	\centering
	\includegraphics[width=1\textwidth]{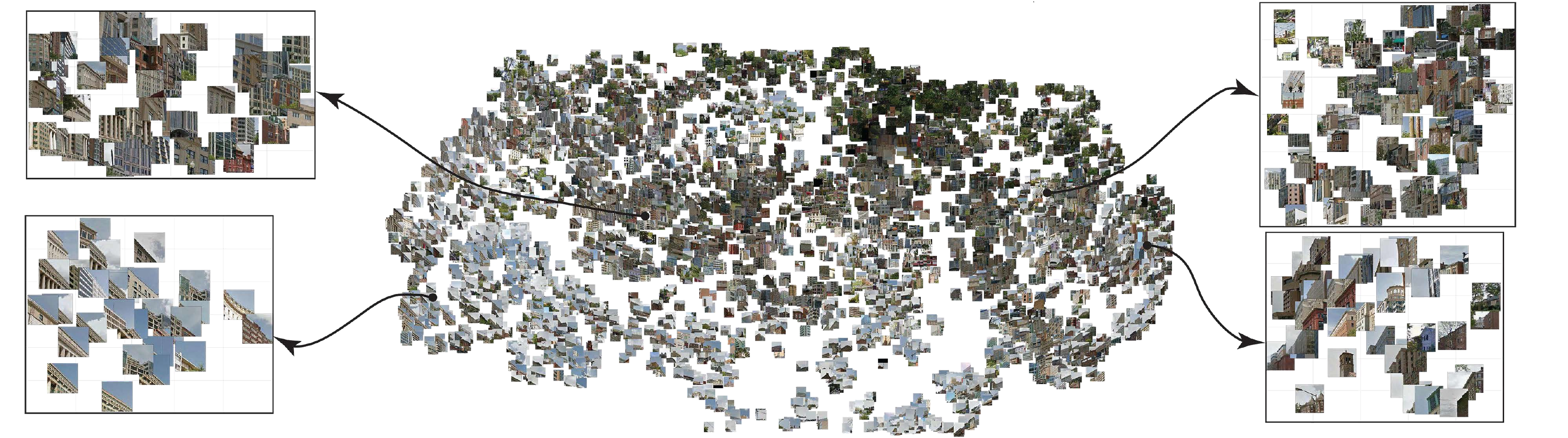}
	\caption{\scriptsize{ \textbf{2D embedding of our representation on 3,000 unseen patches using tSNE}. An organization based on the Manhattan pose of the patches can be seen. See comparable AlexNet's embedding in the {\href{http://3drepresentation.stanford.edu/supplementary_material}{supplementary material}}'s section 6. (best seen on screen)\vspace{-10pt}}}
	\label{fig:wild_pose}
\end{figure*}

\subsubsection{Surface Normals and Vanishing Points}
\textbf{\\}
figure~\ref{fig:wild_pose} shows tSNE embedding of 3,000 unseen patches showing that the organization of the representation space is based on geometry and not semantics/appearance. The ConvNet was trained to estimate the pose between \emph{matching} patches only while in the embedding, the \emph{non-matching} patches with a similar pose are placed nearby. This suggests the representation has generalized the concept of pose to non-matching patches. This indeed has relations to surface normals as the relative pose between an arbitrary and a frontal patch is equal to the pose of the arbitrary patch; figure~\ref{fig:wild_pose} can be perceived as the organization of the patches based on their surface normals.

\begin{figure*}
	\centering\vspace{-20pt}
	\includegraphics[width=1\textwidth]{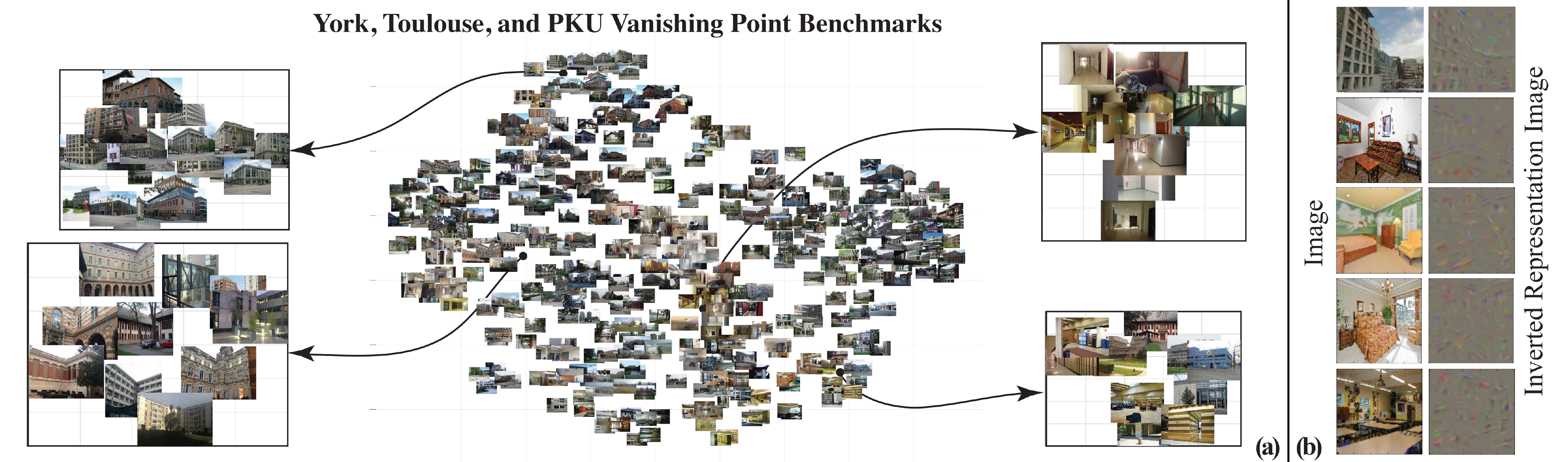}
	\caption{\scriptsize{\textbf{a)} tSNE of a superset of various vanishing point benchmarks~\cite{york,toulouse,pku} (to battle the small size of datasets). \textbf{b)} inversion~\cite{mahendran2015understanding} of our representation. Both plots shows traits of vanishing points. \vspace{-20pt}}}
	\label{fig:vanish_pnts}
\end{figure*}

To better understand how this was achieved, we visualized the activations of the ConvNet at different layers. Similar to other ConvNets, the first few layers formed general gradient based filters while in higher layers, the edges parallel in the physical world seemed to persist and cluster together. This is similar to the concept of vanishing points, and from the theoretical perspective, would be intriguing and explain the pose estimation results, since three common vanishing points are theoretically enough for a full angular pose estimation~\cite{vanish_calib,zissermanBook}. To further investigate this, we generated the inversion of our representation using the method of~\cite{mahendran2015understanding} (see figure~\ref{fig:vanish_pnts}-(b)), which show patterns correlating with the vanishing points of the image. Figure~\ref{fig:vanish_pnts}-(a) also illustrates the tSNE of a superset of several vanishing point benchmarks showing that images with similar vanishing points are embedded nearby. Therefore, we speculate that the ConvNet has developed a representation based on the concept of vanishing points\footnote{\scriptsize{We attempted to quantitatively evaluate this, but the largest vanishing point datasets (e.g., York~\cite{york} and PKU~\cite{pku}) include only 102-200 images for both training and testing. Given a 500D descriptor, it was not feasible to provide a statistically significant evidence.}}. This would also explain the results shown in the following sections. 

\textbf{Surface normal estimation on NYUv2~\cite{silberman2012indoor}:} numerical evaluation on unsupervised surface normal estimation provided in {\href{http://3drepresentation.stanford.edu/supplementary_material}{supplementary material}} sec.~4.


 \begin{figure*}
 	\centering
 	\vspace{0pt}
 	\includegraphics[width=1\textwidth]{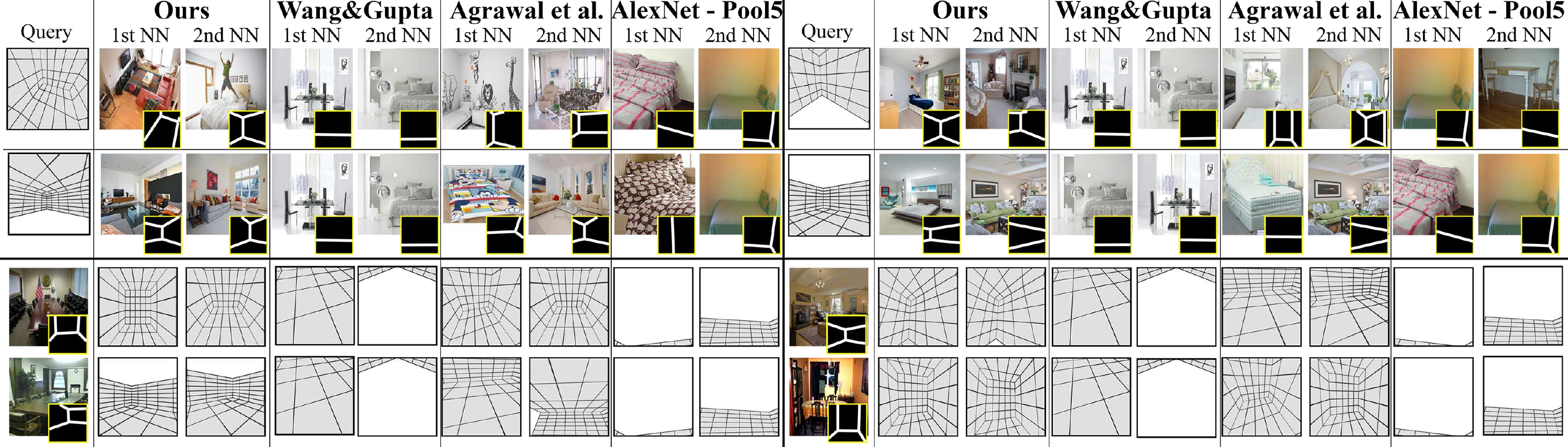}
 	\caption{\scriptsize{\textbf{Scene layout NN search results between LSUN images and synthetic concave cubes defining abstract 3D layouts}. Images with yellow boundary show the ground truth layout.\vspace{-25pt}}}
 	\label{fig:layout}
 \end{figure*}
 
\subsubsection{Scene Layout Estimation}
\textbf{\\}We evaluated our representation on LSUN~\cite{lsun} layout benchmark using the standard protocol~\cite{lsun}. Table~\ref{tab:Layout-est} provides the results of layout estimation using a simple NN classifier on our representation along with two supervised baselines, showing that our representation (with no fine-tuning) achieved a performance close to Hedau et al.'s~\cite{hedau2009recovering} supervised method on this novel task. Table~\ref{tab:Layout-Class} provides the results of layout classification~\cite{lsun} using NN classifier on our representation compared to AlexNets FC7 and Pool5.

\begin{table*}
	\label{tab:joint}
	\parbox{.30\linewidth}{
		\centering
		\vspace{0pt}
		\caption{\scriptsize{Layout Classification (LSUN)}}
		\vspace{0pt}
		\scalebox{0.65}{
			\begin{tabular}{c|c}
				Representation & Classification Accuracy \\
				\hline
				AlexNet FC7 & 45.9\%\\
				AlexNet Pool5 & 47.7\%\\ \hline
				Ours & {57.6}\% \\
			\end{tabular}
			\label{tab:Layout-Class}}
	}
	\hfill \hspace{-0.3mm}
	\parbox{.30\linewidth}{
		\vspace{0pt}
		\centering
		\caption{\scriptsize{Layout Estimation (LSUN)}}
			\vspace{0pt}
		\scalebox{0.65}{
			\begin{tabular}{c|c|c}
				\multirow{2}{*}{Method} & Corner & Pixelwise  \\
				& Error & Error  \\
				\hline
				UIUC (supervised)& 0.11 & 0.17 \\
				Hedau et al. (supervised) & 0.15 & 0.24 \\ \hline
				Ours (unsupervised)& 0.16 & 0.29\\
			\end{tabular}
			\label{tab:Layout-est}}
	}
	\hfill \hspace{-0.3mm}
	\parbox{.30\linewidth}{
		\vspace{0pt}
		\caption{\scriptsize{Object Pose Estimation (PASCAL3D)}}
		\vspace{0pt}
		\centering
		\scalebox{0.65}{
			\begin{tabular}{c|c}
				Method & Av. Pose Error ($\degree$)\\ \hline
				scratch & $34\degree$\\
				AlexNet \tiny{(ImaneNet)}  & \multirow{1}{*}{$23\degree$}\\
				\hline
				Ours & $26\degree$\\
			\end{tabular}}
		}\vspace{0pt}
	\end{table*}

\textbf{Abstraction: Cube$\leftrightarrows$Layout}: To evaluate the abstract generalization abilities of our representation, we generated a sparse set of 88 images showing the interior of a simple synthetic cube parametrized over different view angles. The rendered images can be seen as an abstract cubic layout of a room. We then performed NN search between these images and LSUN dataset using our representations and several baselines. As apparent in figure~\ref{fig:layout}, our representation retrieves meaningful NNs while the baselines mostly overfit to appearance and retrieve either an incorrect or always the same NN. This suggests our representation could abstract away the irrelevant information and encode some information essential to the 3D of the image.

\begin{figure*}
	\centering
	\includegraphics[width=1\textwidth]{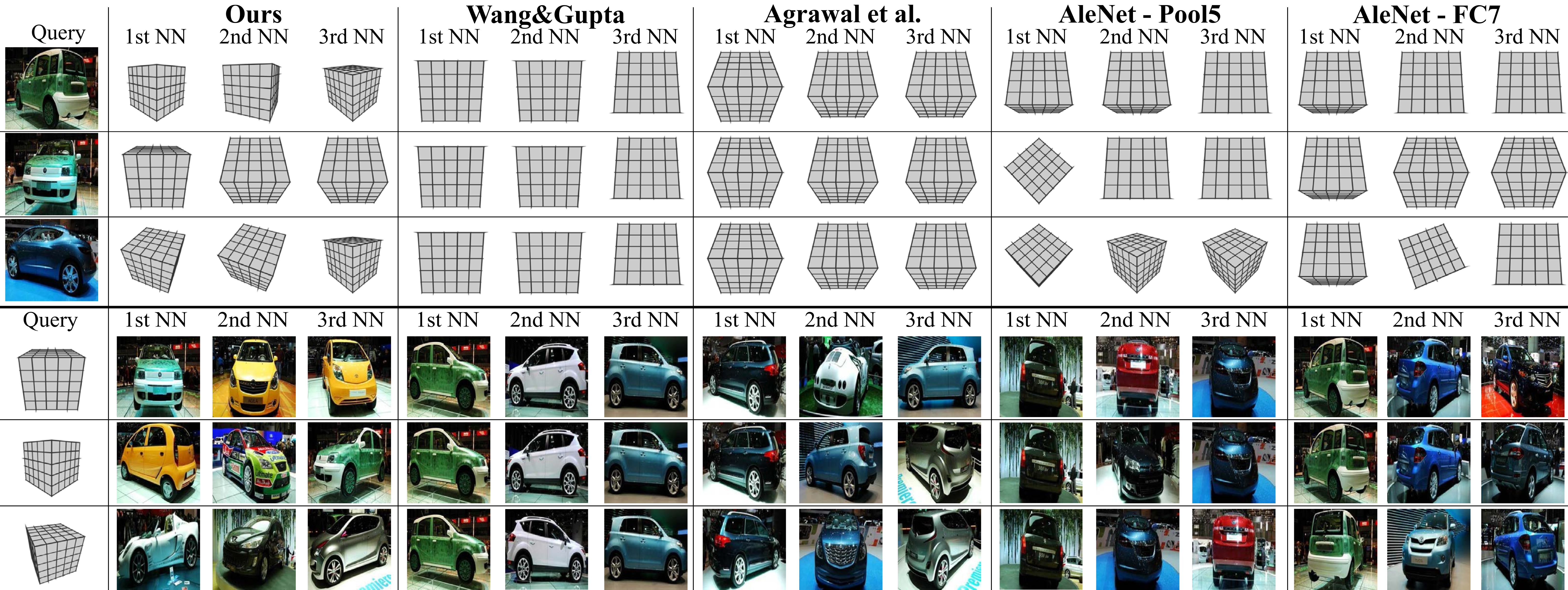}
	\caption{\scriptsize{\textbf{NN search results between EPFL dataset images and a synthetic cube defining an abstract 3D pose.} See the {\href{http://3drepresentation.stanford.edu/supplementary_material}{supplementary material}} (section 5) for tSNE embedding of all cubes and car poses in a joint space. Note that the 3D poses defined by the cubes are 90\degree congruent.  \vspace{-10pt}}}
	\label{fig:EPFL}
\end{figure*}


\vspace{10pt}
\subsubsection{3D Object Pose Estimation}
$\\$

\textbf{Abstraction: Cube$\leftrightarrows$Object}: we performed a similar abstraction test between a set of 88 convex cubes and the images of EPFL Multi-View Car dataset~\cite{epfl}, which includes a dense sampling of various viewpoints of cars in an exhibition. We picked this simple cube pattern as it is the simplest geometric element that defines three vanishing points. The same observation as the abstraction experiment on LSUN's is made here with our NNs being meaningful while baselines mostly overfit to appearance with no clear geometric abstraction trait(figure~\ref{fig:EPFL}).

\begin{figure*}
	\centering
	\includegraphics[width=1\textwidth]{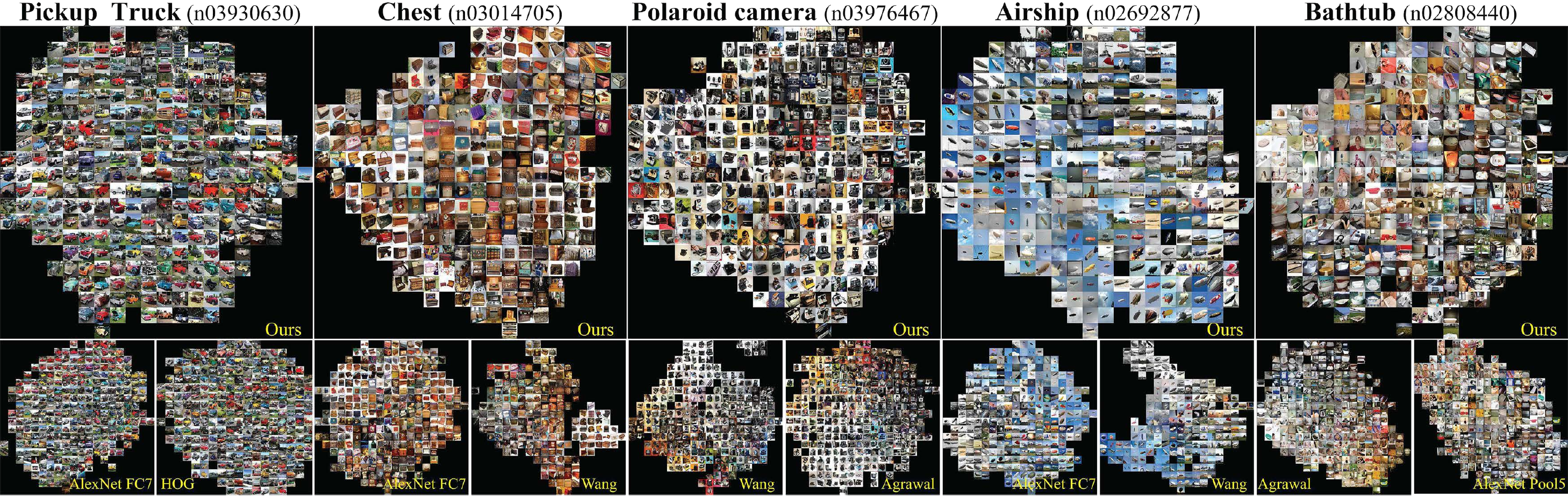}
	\caption{\scriptsize{\textbf{tSNE of several ImageNet categories using our unsupervised representation} along with several baselines. Our representation manifests a meaningful geometric organization of objects. tSNE of more categories in the {\href{http://3drepresentation.stanford.edu/supplementary_material}{supplementary material}} and the {\href{http://3Drepresentation.stanford.edu/}{website}}. (best seen on screen)\vspace{0pt}}}
	\label{fig:imagenet}
\end{figure*}

\textbf{ImageNet}: Figure~\ref{fig:imagenet} shows the tSNE embedding of several ImageNet categories based on our representation and the baselines. The embeddings of our representation are geometrically meaningful, while the baselines either perform a semantic organization or overfit to other aspects, such as color. 

\begin{figure*}
	\centering
	\vspace{0pt}
	\includegraphics[width=1\textwidth]{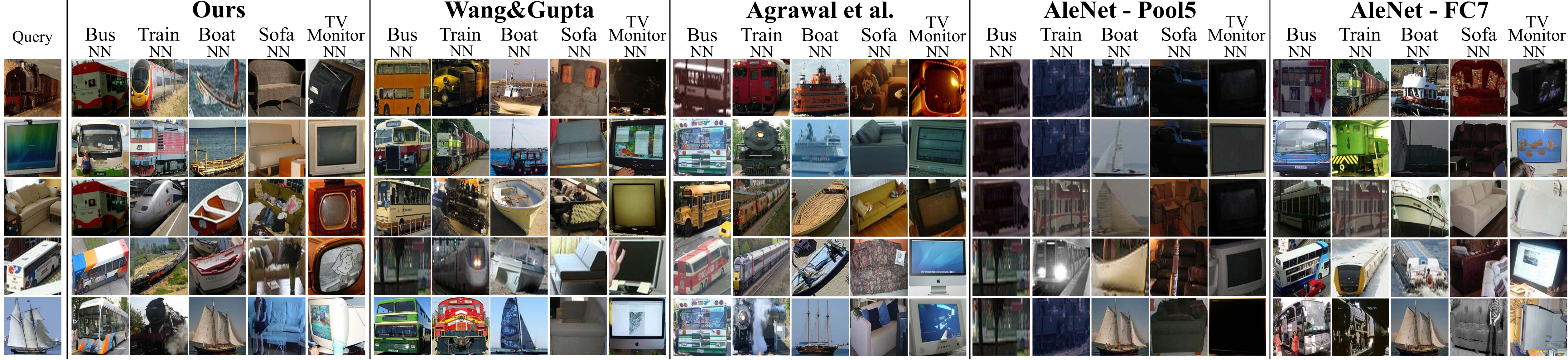}
	\caption{\scriptsize{\textbf{Qualitative results of cross-category NN-search on PASCAL3D} using our representation along with baselines.}\vspace{0pt}}
	\label{fig:pascal3d}
\end{figure*}

\textbf{PASCAL3D}:
Figure~\ref{fig:pascal3d} shows cross-category NN search results for our representation along with several baselines. This experiment also evaluates a certain level of abstraction as some of the object categories can be drastically different looking. We also quantitatively evaluated on 3D object pose estimation on PASCAL3D. For this experiment, we trained a ConvNet from scratch, fine-tuned AlexNet pre-trained on ImageNet, and fine-tuned our network; we read the pose out using a linear regressor layer.\footnote{\scriptsize{The classes of boat, sofa, and chair were showing a performance near statistically informed random for all methods and were removed from the evaluations.}} Our results outperform scratch network and come close to AlexNet that has seen thousands of images from the same categories from ImageNet and other objects. Note that certain aspects of object pose estimation, e.g., distinguishing between the front and back of a bus, are more of a semantic task rather than geometric/3D. This explains a considerable part of the failures of our representation which is object/semantic agnostic.

\vspace{-20pt}
\section{Discussion and Conclusion}
\label{sec:conclusion}
To summarize, we developed a generic 3D representation through solving a set of supervised foundational proxy tasks. We reported state-of-the-art results on the supervised tasks and showed the learned representation manifests generalization and abstraction traits. However, a number of questions remain open: 

Though we were inspired by cognitive studies in defining the foundational supervised tasks leading to a generalizable representation, this remains at an inspiration level. Given that a `taxonomy' among basic 3D tasks has not been developed, it is not concretely defined which tasks are foundational and which ones are secondary. Developing such a taxonomy (i.e., whether task A is inclusive of, overlapping with, or disjoint from task B) or generally efforts understanding the task space would be a rewarding step towards soundly developing the \emph{3D complete} representation. Also, semantic and 3D aspects of the visual world are tangled together. So far, we have developed independent semantic and 3D representations, but investigating concrete techniques for integrating them (beyond simplistic late fusion or ConvNet fine-tuning) is a worthwhile future direction for research. Perhaps, inspirations from partitions of visual cortex could be insightful towards developing the ultimate \emph{vision complete} representation.

\noindent\textbf{Acknowledgement:} We gratefully acknowledge the support of ICME/NVIDIA Award (1196793-1-GWMUE), MURI (1186514-1-TBCJE), and Nissan (1188371-1-UDARQ).

\clearpage

\bibliographystyle{splncs03}
\bibliography{representation_bib}

\end{document}